\documentclass[letterpaper]{article} 
\usepackage{aaai23}  
\usepackage{times}  
\usepackage{helvet}  
\usepackage{courier}  
\usepackage[hyphens]{url}  
\usepackage{graphicx} 
\usepackage{natbib}  
\usepackage{caption} 
\usepackage{amsmath}
\usepackage{amsthm}
\usepackage{tabularx}
\usepackage{algorithm}
\usepackage{algorithmic}
\usepackage{booktabs}
\usepackage{siunitx}
\usepackage{multirow}
\usepackage{algorithm}
\usepackage{algorithmic}

\usepackage{newfloat}
\usepackage{listings}
\DeclareCaptionStyle{ruled}{labelfont=normalfont,labelsep=colon,strut=off} 
\floatstyle{ruled}
\newfloat{listing}{tb}{lst}{}
\floatname{listing}{Listing}

\title{Mitigating Negative Transfer in Multi-Task Learning \\ with Exponential Moving Average Loss Weighting Strategies \\ (Abstract and Appendix)}

\author {
    Anish Lakkapragada,\textsuperscript{\rm 1}
    Essam Sleiman,\textsuperscript{\rm 3}
    Saimourya Surabhi,\textsuperscript{\rm 1,2, *}  
    Dennis P. Wall\textsuperscript{\rm 1,2}
}
\affiliations {
    \textsuperscript{\rm 1} Department of Biomedical Data Science, Stanford University, Stanford CA 94305, \\
    \textsuperscript{\rm 2} Department of Pediatrics, Stanford University, Stanford CA 94305, \\
    \textsuperscript{\rm 3} Department of Computer Science, University of California, Davis CA 95616 \\
    \{anishlk, mourya, dpwall\}@stanford.edu,
    esleiman@ucdavis.edu \\
    \textsuperscript{\rm *}Corresponding Author: mourya@stanford.edu
}

\begin{document}

\maketitle

\begin{abstract}

Multi-Task Learning (MTL) is a growing subject of interest in deep learning, due to its ability to train models more efficiently on multiple tasks compared to using a group of conventional single-task models. However, MTL can be impractical as certain tasks can dominate training and hurt performance in others, thus making some tasks perform better in a single-task model compared to a multi-task one. Such problems are broadly classified as negative transfer, and many prior approaches in the literature have been made to mitigate these issues. One such current approach to alleviate negative transfer is to weight each of the losses so that they are on the same scale. Whereas current loss balancing approaches rely on either optimization or complex numerical analysis, none directly scale the losses based on their observed magnitudes. We propose multiple techniques for loss balancing based on scaling by the exponential moving average and benchmark them against current best-performing methods on three established datasets. On these datasets, they achieve comparable, if not higher, performance compared to current best-performing methods.

\end{abstract}

\section{Introduction}


A plethora of loss balancing methods have emerged to prevent certain tasks in MTL from dominating at the expense of other tasks' performances in training. In contrast to current loss balancing methods that employ complex optimization or numerical analysis, we propose a much simpler method, directly scaling each loss by its observed magnitude. To accomplish this, we simply scale task losses by their exponential moving averages (EMAs). We merge other loss balancing techniques centered around weighting tasks based on their training rates with ours. Our techniques are competitive and superior to several other best-performing techniques on the CelebA, AffWild2, and AffectNet datasets.

\section{Methods}

Current loss balancing methods generally either learn the loss coefficients for each task through parameter optimization or derive them from some measure of a training rate. GradNorm \cite{chen2018gradnorm} employs both methods in determining the loss coefficients by optimizing them to keep the gradient magnitudes for all tasks more or less equivalent depending on their individual training rates. Dynamic Weight Average or DWA \cite{liu2019end} calculates the descending speed of convergence of a task based on the ratio of that task's loss at the current compared to the past training iteration and assigns tasks with a higher ratio (and thus slower convergence) greater loss weightage. However, DWA doesn't account for the actual magnitudes of the individual task losses. 

No method we have seen so far, to the best of our knowledge, considers simply scaling each of the task losses by their magnitudes. We accomplish this by dividing each of the task losses by their observed exponential moving average (EMA) so they are all on the same scale of one. 

\begin{equation} \label{eq:1}
\begin{split}
    \tilde{L}_{k}(t) = \beta L_{k}(t) + (1-\beta) \tilde{L}_{k}(t-1)\\
    \mathcal{L}_{MTL}(t) = \sum_{k=1}^{K} \lambda_{k}(t) L_{k}(t), \lambda_{k}(t)= \dfrac{1}{\tilde{L}_{k}(t)}
\end{split}
\end{equation}

\begin{table*}[t]
    \centering
    \footnotesize
    \begin{tabularx}{0.7\textwidth}{@{}llllllll@{}}
    \toprule
    \textbf{Method} & \textbf{Baseline} & \textbf{GradNorm} & \textbf{UW} & \textbf{DWA} & \textbf{EMA} & \textbf{DWEMA} & \textbf{REMA} \\
    \midrule
    \textbf{Performance} & 0.772 & 0.771 & 0.768 & 0.772 & 0.782 & 0.788 & \textbf{0.788} \\
    \bottomrule
    \end{tabularx}
    \caption{Overall performance on the 40 CelebA tasks for our experiments.}
    \label{table:celeba}
\end{table*}

We show the formulation for our proposed method in Equation \ref{eq:1}, where $t$ is the training iteration and $\beta$ is the weight term in calculating the task loss EMA $\tilde{L}_{k}$ -- the reciprocal of which is the loss coefficient $\lambda_{k}$ for task $k$. \citeauthor{chen2018gradnorm} notes that the Uncertainty Weighting technique \cite{kendall2018multi}, a successful loss balancing technique which learns $\lambda_{k}$ to optimize the overall loss through gradient descent, often learns $\lambda_{k}$ $\approx$ $1/L_{k}$. This validates our idea to directly scale task losses by their moving average. However, they also note Uncertainty Weighting (UW) can lead to volatile spikes in the $\lambda_{k}$, which leads to performance deterioration -- in our case, $\beta$ is a hyperparameter to help prevent such issues by tuning how fast to adapt the task loss EMA and thus the loss coefficients themselves.

We also merge our idea with DWA in our Rated Exponential Moving Average (REMA) method, where each of the losses are first scaled by their EMA and then shifted accordingly by their training rates $r_{k}$ for task $k$ to increase loss weightage on tasks with slower convergence. This is shown in Equation \ref{eq:2}.

\begin{equation}
    \label{eq:2}
    \mathcal{L}_{MTL}(t) = \sum_{k=1}^{K} r_{k}(t) \dfrac{L_{k}(t)}{\tilde{L}_k(t)}, r_{k}(t) = \dfrac{L_{k}(t-1)}{L_{k}(t-2)} 
\end{equation}

We compare this to directly scaling the DWA coefficients by $\tilde{L}_{k}$, instead of just $r_{k}$. We refer to this method as Dynamic Weighted Exponential Moving Average (DWEMA). Unlike DWA, both DWEMA and REMA actively prevent task domination by first scaling losses by their magnitudes before applying training rate adjustments. 
    
\section{Experiments} 

We test our models on three datasets, CelebA \cite{liu2015faceattributes}, AffWild2 \cite{kollias2022abaw}, and AffectNet \cite{mollahosseini2017affectnet}. All contain cropped and aligned facial images; we define the tasks for each of the datasets as all the attributes labeled for each image (except for the binary labels for each of the twelve action units annotated in AffWild2, which are treated as one task). Our overall performance metrics are determined by prior metrics for these datasets or a standard average of F1 for the 40 tasks in the CelebA dataset.


Our AffWild2 and AffectNet MTL models were trained using standard hard parameter sharing on a ResNet50 backbone. On the CelebA dataset, due to computational constraints, we use a ResNet18 backbone feeding to a dense layer of 40 neurons to predict on the 40 tasks. In order to frame our CelebA experiments as MTL, we backpropagate on each task individually, based on its individual loss coefficient.


We train all our models with the same initialization of parameters, with $\beta$ to 0.2 on CelebA runs and 0.1 for AffWild2 and AffectNet runs. The performances for CelebA are detailed in Table \ref{table:celeba} and the individual task and overall performances for AffWild2 and AffectNet are described in Table \ref{table:aaresults}. 

\begin{table}[ht]
    \centering
    \scalebox{0.9}{
    {
        {\begin{tabularx}{0.53
    \textwidth}{lllllll}
    \toprule
    \textbf{Method} & \textbf{Dataset} & \textbf{AU} & \textbf{Emotion} & \textbf{VA} & \textbf{Overall} \\
    \midrule
    \addlinespace
    \multirow{2}{-0.4ex}{\textbf{Single Task}} & \textbf{AffWild2} & 0.574 & 0.057 & 0.068 & 0.699 \\
    \addlinespace
    & \textbf{AffectNet} & -- & 0.426 & 0.419  & 0.843  \\
    \hline
    \addlinespace
    \multirow{2}{*}{\textbf{Baseline}} & \textbf{AffWild2} & 0.579 & \textbf{0.082} & 0.083 & 0.744 \\
    \addlinespace
    & \textbf{AffectNet} & -- & 0.425 & 0.428 & 0.853 \\
    \hline 
    \addlinespace
    \multirow{2}{*}{\textbf{GradNorm}} & \textbf{AffWild2} & 0.579 & 0.080 & 0.072 & 0.730 \\
    \addlinespace
    & \textbf{AffectNet} & -- & 0.401 & 0.408 & 0.809 \\
    \hline
    \addlinespace
    \multirow{2}{*}{\textbf{UW}} & \textbf{AffWild2} &0.578 & 0.081 & 0.087 & 0.746 \\
    \addlinespace
    & \textbf{AffectNet} & -- & 0.393 & 0.410 & 0.803 \\
    \hline
    \addlinespace
    \multirow{2}{*}{\textbf{DWA}} & \textbf{AffWild2} & 0.580 & 0.079 & 0.102 & 0.760 \\
    \addlinespace
    & \textbf{AffectNet} & -- & 0.407 & 0.406 & 0.813 \\
    \hline
    \addlinespace
    \multirow{2}{*}{\textbf{REMA}} & \textbf{AffWild2} & 0.586 & 0.080 & 0.100 & 0.764 \\
    \addlinespace
    & \textbf{AffectNet} & -- & 0.421 & 0.449 & 0.869 \\
    \hline
    \addlinespace
    \multirow{2}{*}{\textbf{DWEMA}} & \textbf{AffWild2} &0.588 & 0.081 & 0.106 & 0.775 \\
    \addlinespace
    & \textbf{AffectNet} & -- & 0.425 & 0.443 & 0.868 \\
    \hline
    \addlinespace
    \multirow{2}{*}{\textbf{EMA}} & \textbf{Affwild2} &\textbf{0.590} & 0.080 &
    \textbf{0.133} & \textbf{0.800} \\
    \addlinespace
     & \textbf{AffectNet} & -- & \textbf{0.427} & \textbf{0.471} & \textbf{0.898} \\
    \bottomrule
    \end{tabularx}
    }
    }
    }
    \caption{Analysis of each of the performances on the Action Units (AUs, only for AffWild2), Emotion, and Valence \& Arousal (VA) tasks and overall performance for our AffWild2 and AffectNet experiments.}
    \label{table:aaresults}
\end{table}

We observe the merit of scaling losses by their EMA through the higher performance of the EMA and variants, REMA and DWEMA, approaches compared to other past methods. These two variants based on training rates have a bigger impact on performance in the CelebA dataset compared to the others. We conjecture that when training rates $r_{k}$ are more varied, weighting certain tasks by their training rate would be more beneficial for performance through a comparison on the training rates for each dataset. 





\section{Conclusion}
 
We propose three EMA-based techniques for mitigating negative transfer in MTL models which achieve superior performance on the CelebA, AffectNet, and AffWild2 dataset compared to several best-performing methods. Overall, we reason that our method's higher performance is due to the explicit scaling of the losses being more defined than in past approaches. We foresee our merge of MTL with expression-cented AI efforts to be more relevant in the future as MTL provides more efficient computing on edge devices. 



\section{Acknowledgements}

This work was supported in part by funds to DPW from the NIH, the NSF, MediaX, WuTsai Neurosciences Institute.

\newpage

\section{APPENDIX}

We present further information on our implementation, hyperparameter choices, and reasoning in our abstract in this supplementary material.

\section{A. Implementation Details}

\subsection{A.1 CelebA}

CelebA \cite{liu2015faceattributes} is a dataset containing around 200,000 cropped and aligned facial images. Each image is annotated with 40 binary attributes describing the face; example attributes include whether the person has eyeglasses, is wearing a hat, is smiling, etc.

\begin{figure}[htp]
    \centering
    \includegraphics[width=9cm]{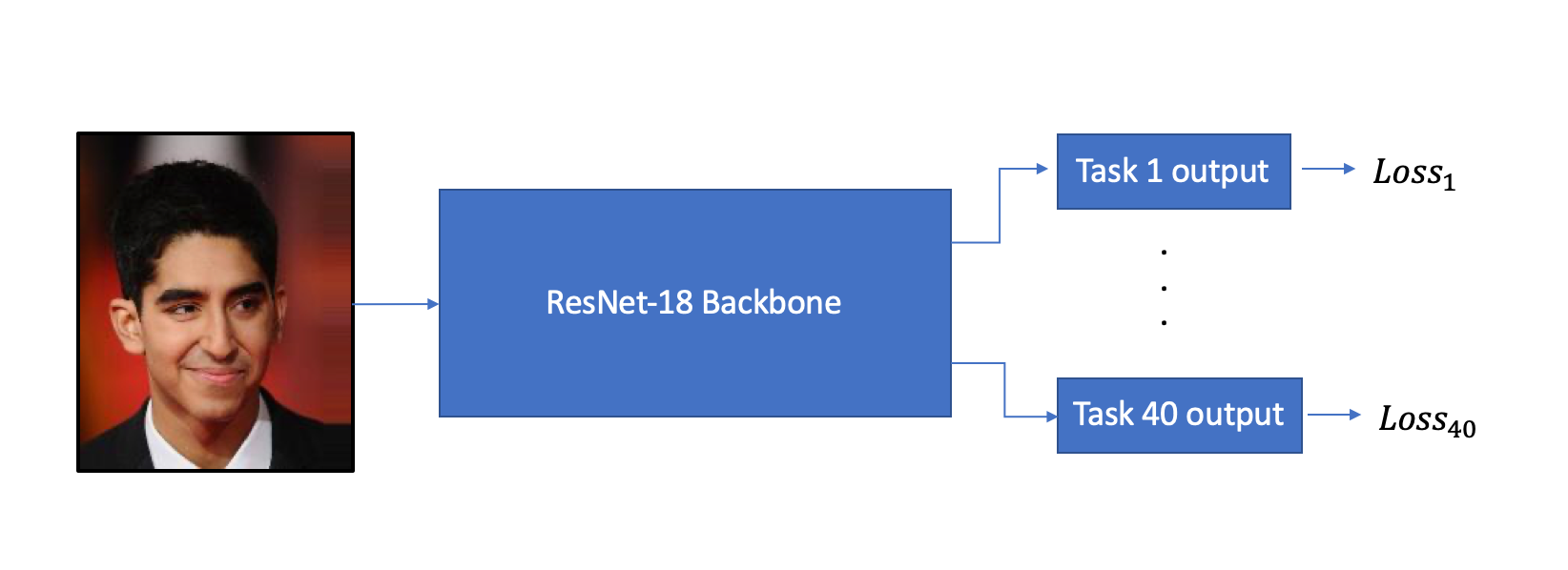}
    \caption{CelebA MTL architecture used in our experiments.}
    \label{fig:celebamodel}
\end{figure}

For our multi-task learning (MTL) experiments on the CelebA dataset, we utilize a ResNet18 \cite{he2016deep} backbone and take its 1000-dimensional output vector and feed it into a fully connected layer that maps 1000 input neurons to 40 output units. Each of these neurons yields a binary value for that particular task; thus we optimize each task individually with the binary cross-entropy loss function. To frame this model as an MTL problem, we apply the loss function to each task individually and scale each of the 40 tasks' losses based on their respective loss coefficients before backpropagation. For clarity, a diagram of our model's architecture is shown in Figure \ref{fig:celebamodel}. 

Before training this model, we resize each image in the CelebA dataset to a size of 64 x 64 x 3 and apply image normalization with a mean and standard deviation of $0.5$ in each channel. We train these models with a learning rate of 1e-04 using Adam optimizer \cite{kingma2014adam}. 

\subsection{A.2 AffWild2 and AffectNet}

\subsubsection{AffWild2}

We evaluate our techniques on the AffWild2 dataset \cite{kollias2022abaw, kollias2020va}. AffWild2 contains cropped and aligned facial images, annotated with the subject's categorical emotion, the binary presence of each of the 12 action units (AUs) in the subject, and valence and arousal (VA) values. Valence and arousal are continuous values; the valence dimension ranges from highly negative to highly positive expressions and the arousal dimension ranges from calm to excited or agitated expressions \cite{kensinger2004remembering}.  We use the publicly available training and validation data from the ABAW 2022 Multi-Task Learning Challenge \cite{kollias2022abaw} as our respective training and testing data. In total, we use approximately 142,400 images for training and 27,000 images for testing.

\begin{figure}[htp]

    \includegraphics[width=9cm]{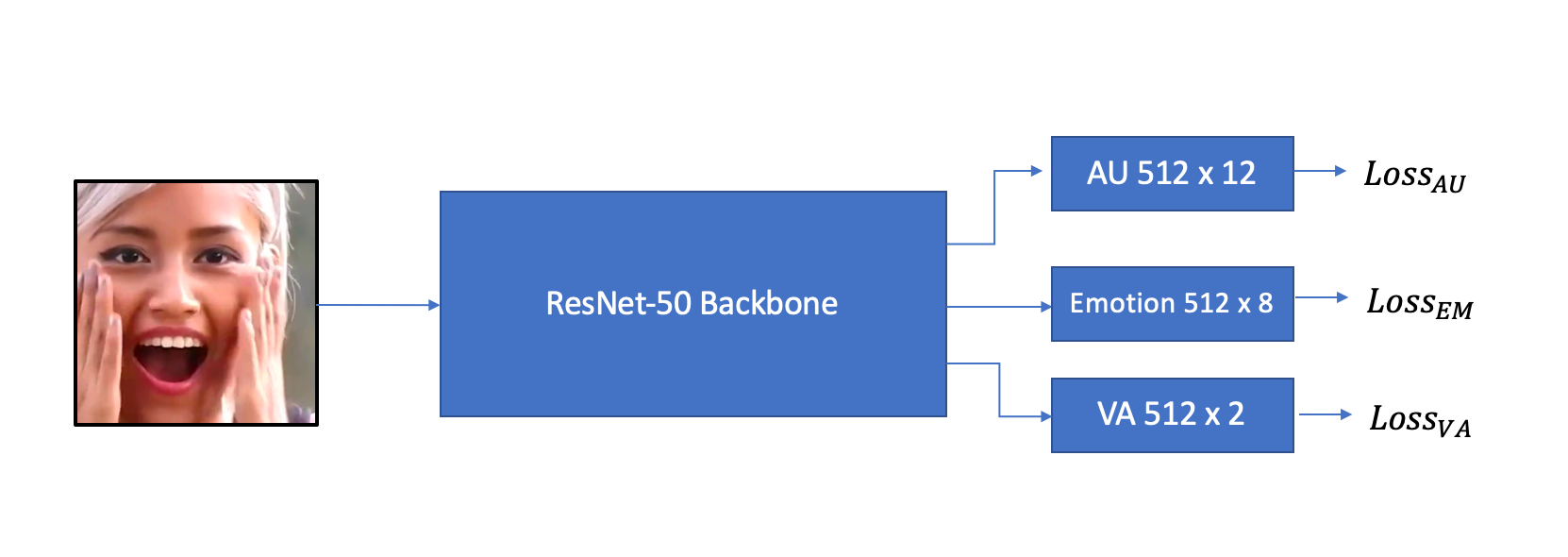}
    \caption{AffWild2 MTL architecture used in our experiments.}
    \label{fig:affwild-model}
\end{figure}

The backbone of our AffWild2 MTL model is a ResNet-50 pretrained on VGGFace2 -- this backbone's output representation is then fed to separate neural networks for each task. Each of these networks consist of one dense layer of 512 neurons, that are fed into a ReLU activation function before being fed to a final dense layer with a specific number of neurons and activation function based on the specific task. We choose this configuration based on a prior model in \cite{thinh2021emotion} that trained on AU and emotion tasks on the AffWild2 dataset. A diagram of this architecture is shown in Figure \ref{fig:affwild-model}. We found that training all MTL and single-task learning (STL) models with an Adam optimizer at a learning rate of 5e-06 performed best.

\subsubsection{AffectNet}

The final dataset we evaluate our techniques on is the AffectNet dataset \cite{mollahosseini2017affectnet}. AffectNet is one of the largest datasets containing expression, valence, and arousal annotations in facial images. We train our models with the provided AffectNet subset of almost 300,000 images, with their original size of 256 x 256 x 3. We apply the same image normalization as with the CelebA dataset, and train our models with the Adam optimizer at a learning rate of 1e-05. Our AffectNet models use the same architecture as the AffWild2 MTL model, however with the action units network removed. 

\section{B. Performance Evaluation} 

For comparing models as a whole, we employ dataset-specific metrics to calculate the model's overall performance. For the CelebA dataset, we use the mean F1-score of all the 40 binary tasks to calculate the overall performance of the model.

For AffWild2 experiments, we follow the same performance metric used in the ABAW 2022 Multi-Task Learning Challenge. For our experiments, we report the individual task performance on Action Units (AUs) and Emotion tasks as the F1-score between each task's predictions and labels. For assessing valence and arousal performance, we take the concordance correlation coefficient (CCC) between each of these output's predictions and labels. The sum of the F1-scores on both the AU and Emotion tasks is added with the average CCC between the predictions and labels for valence and arousal outputs to give the overall performance. In equation form, the overall performance can be described as: $0.5 * (CCC_{V} + CCC_{A}) + (\sum_{i=1}^{12} F1_{i}^{AU} / 12) + F1_{Emotion}$. Our overall performance metric for AffectNet is the same as for AffWild2 without the action units average F1 in the summation.

\section{C. Hyperparameter Choices} 

\subsection{C.1 DWA}
Dynamic Weight Average (DWA) \cite{liu2019end} is one of the techniques we draw inspiration from and benchmark against. DWA contains a hyperparameter on the temperature, $T$, which measures how similar each of the loss coefficients should be. Higher values of $T$ lead to closer loss coefficient values. We experiment on three different values of $T$, including $T$ equal to 2.0 (the suggested value), and show their overall performances on our three datasets in  Table \ref{table:tempDWA}. 

\begin{table}[ht]
    \centering
    \begin{tabular}{|c|c|c|c|}
        \hline
        Temperature & CelebA & AffWild2 & AffectNet \\
        \hline
        $T$=2 & 0.767 & 0.741 & \textbf{0.813} \\
        \hline
        $T$=1 & 0.769 & 0.727 & 0.081  \\
        \hline
        $T$=0.5 & \textbf{0.772} & \textbf{0.76} & 0.805 \\
    \hline
    \end{tabular}
    \caption{Effect of temperature $T$ on overall performance in DWA, evaluated on all our datasets.}
    \label{table:tempDWA}
\end{table}

We report and compare results for DWA and Dynamic Weight Exponential Moving Average (DWEMA) with $T$ set to 0.5 on the CelebA and AffWild2 datasets and $T$ set to 2.0 on the AffectNet dataset as these values lead to the highest observed performance.

\subsection{C.2 GradNorm}
Another method we benchmark our techniques against, GradNorm \cite{chen2018gradnorm}, has a parameter $\alpha$ which dictates the degree to which the tasks should be training at the same rate when calculating the optimal loss coefficients. Higher values of $\alpha$ force each of the tasks to train at a closer rate of convergence. We notice that performance for different values of $\alpha$ fluctuates by small amounts on differing parameter initialization of the model, thus we default to the suggested value of $\alpha$ in the paper, $1.5$. GradNorm also requires a subset of the weights in the model to observe gradient magnitudes to calculate the loss coefficients; we set the last shared convolutional layer's parameters as this subset of weights (as recommended in the paper.)

\subsection{C.3 EMA Methods (Ours)}

\begin{figure}[htp]
    \centering
    \includegraphics[width=8cm]{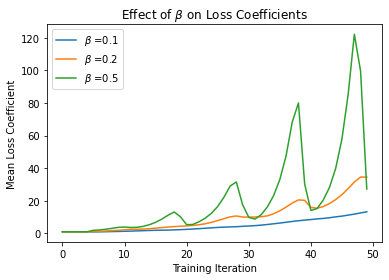}
    \caption{Impact of $\beta$ on the progression of the mean of the loss coefficients during training on the AffWild2 dataset.}
    \label{fig:losscoeff}
\end{figure}

In our methods to scale task losses based on their exponential moving averages (EMAs), we tune the weight parameter $\beta$ in our method. As stated in our abstract, $\beta$ controls the rate that these EMAs, and therefore the loss coefficients themselves, adapt to updates in the task loss values. We show $\beta$'s affect on the loss coefficients in Figure \ref{fig:losscoeff}, which displays the mean of the loss coefficients during training on the AffWild2 dataset for three $\beta$ values. The effect of $\beta$ on overall performance for all our datasets is detailed in Table \ref{table:betaPerf}.

\begin{table}[ht]
    \centering
    \begin{tabular}{|c|c|c|c|}
        \hline
        $\beta$ Value & CelebA & AffWild2 & AffectNet \\
        \hline
        $\beta = 0.5$ & 0.754 & 0.757 & 0.839 \\
        \hline
        $\beta = 0.2$ & \textbf{0.783} & 0.767 & 0.829  \\
        \hline
        $\beta = 0.1$ & 0.778 & \textbf{0.800} & \textbf{0.898} \\
    \hline
    \end{tabular}
    \caption{Effect of $\beta$ on overall performance on all our evaluated datasets.}
    \label{table:betaPerf}
\end{table}

From the graph, we see lower values of $\beta$ prevent extreme spikes in the loss coefficients due to more gradual updating of the task loss EMA from current iterations. We note this behavior is invariant of the applied dataset. In general, values of $\beta$ equal to $0.2$ or lower outperform higher values of $\beta$ above or equal to $0.5$. For all EMA-based approaches, we set $\beta$ to $0.2$ on the CelebA dataset and $0.1$ on the AffWild2 and AffectNet dataset. 

When experimenting with higher values of $\beta$, we noticed that the sharp spikes in the loss coefficients lead to overall loss typically spiking in the next training iteration. This same phenomena is detailed in the GradNorm paper \cite{chen2018gradnorm} that a related method, Uncertainty Weighting \cite{kendall2018multi}, would learn the loss coefficients but ultimately deteriorate in performance due to similar rapid spikes in loss coefficients. Through the use of tuning $\beta$, we hope to alleviate such issues of past methods. 

\section{D. Further Information} 

\subsection{D.1 Training rates on EMA Variants Comparison}

\begin{figure}[htp]
    \centering
    \includegraphics[width=8cm]{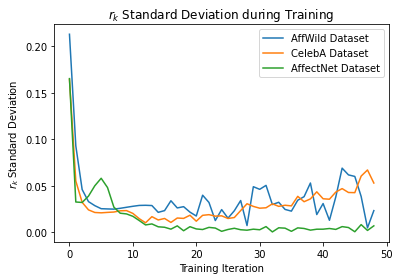}
    \caption{Standard deviation of training rates throughout the MTL baseline model training on all our three datasets.}
    \label{fig:trainingratestd}
\end{figure}

We conjecture that Rated Exponential Moving Average (REMA) and DWEMA led to greater performance gains over the vanilla EMA approach on the CelebA dataset compared to the AffWild2 and AffectNet datasets due to loss weightage based on training rates being more beneficial in cases where the training rates are more varied. We define the (inverse) training rate $r_{k}$ for task $k$ at the training iteration $t$ to be the ratio $\dfrac{L_{k}(t-1)}{L_{k}(t-2)}$ comparing the current loss for a certain task to the loss at the past training iteration. We measure the variation of the training rates at a given iteration as the standard deviation of all the training rates at that time. We plot the standard deviations of the training rates during training on all our baseline MTL models in Figure \ref{fig:trainingratestd}.

The variation of CelebA's training rates consistently increase more throughout training compared to AffWild2, which fluctuates, and AffectNet, which consistently has the lowest amount of variation. Pairing this data with the fact that REMA and DWEMA, which use a task's training rate to adjust each task's loss weighting, led to a greater performance gain on CelebA -- our hypothesis is that weighting tasks based on their training rate is more desirable in cases where the training rates of the task naturally vary more.






\bibliography{aaai23-supp.bib}

\end{document}